# Dynamic Knowledge Graph-based Dialogue Generation with Improved Adversarial Meta-Learning

Hongcai Xu*, Junpeng Bao, Gaojie Zhang

Xi'an Jiaotong University, Xi'an, Shaanxi, 710049, P.R. China

xajt1822@stu.xjtu.edu.cn

## Abstract

Knowledge graph-based dialogue systems are capable of generating more informative responses and can implement sophisticated reasoning mechanisms. However, these models do not take into account the sparseness and incompleteness of knowledge graph (KG), and current dialogue models cannot be applied to dynamic KG. This paper proposes a dynamic Knowledge graph-based dialogue generation method with improved adversarial Meta-Learning (KDAD). KDAD formulates dynamic knowledge triples as a problem of adversarial attack, and incorporates the objective of quickly adapting to dynamic knowledge-aware dialogue generation. We train a knowledge graph-based dialog model with improved ADML using minimal training samples. The model can initialize the parameters and adapt to previous unseen knowledge, so that training can be quickly completed based on only a few knowledge triples. We show that our model significantly outperforms other baselines. We evaluate and demonstrate that our method adapts extremely fast and well to dynamic knowledge graph-based dialogue generation.

## 1. Introduction

Data-driven neural dialogue systems usually learn from a massive amount of human-human conversational corpus using End-to-End (E2E) learning (Sutskever et al., 2014; Serban et al., 2016), without combining hand-crafted rules or templates. However, Sequence-to-sequence (seq2seq) model tend to produce generic or incoherent responses, such as "I don't know". Recently, in order to generating high-quality and informative conversation responses, external knowledge is employed in open-domain dialogue systems, including unstructured texts (Ghazvininejad et al., 2018) or structured knowledge representation (Liu et al., 2018; Young et al., 2018; Zhou et al., 2018a; Liu et al., 2019).

Knowledge graph's entities and edges can narrow the candidate set very quickly. Moreover, knowledge triples can enhance capability of generating informative and diverse conversational responses. Because of the high human annotation cost, a limited number of triples suffer from information insufficiency for response generation. Nonetheless, the model capability of zero-shot adaptation to dynamic knowledge graph has rarely been considered. Entities or relations in dynamic knowledge graphs are temporal and evolve as a single time scale process (Tuan et al., 2019). For example, as illustrated in Figure 1, the entity Jin-Xi was originally related to the entity

Feng, Ruozhao with the type EnemyOf, but then evolved to be related to the entity Nian, Shilan.

In this paper, we propose an improved adversarial meta-learning algorithm (Yin et al., 2018) to facilitate knowledge aware dialogue generation. Adversarial meta-learning is presented based on Model-Agnostic Meta-Learning (MAML) (Finn et al., 2017). MAML is a simple, general, and effective optimization algorithm aiming to learn an internal feature that is broadly applicable to all tasks in a task distribution $p(\mathcal{T})$, rather than a single task. The key idea of this article is considering dynamic entities and relations as adversarial samples, and fully utilizing knowledge graph-based dialog data to learn an initialization which adapt to new knowledge triples quickly. By combining Qadpt (Tuan et al., 2019), a seq2seq neural conversation model with copy mechanism (Xing et al., 2017), we implement the improved ADML algorithm to learn an optimal initialization. We evaluate and show that our model outperforms the state-of-the-art baselines (Qadpt and TAware).

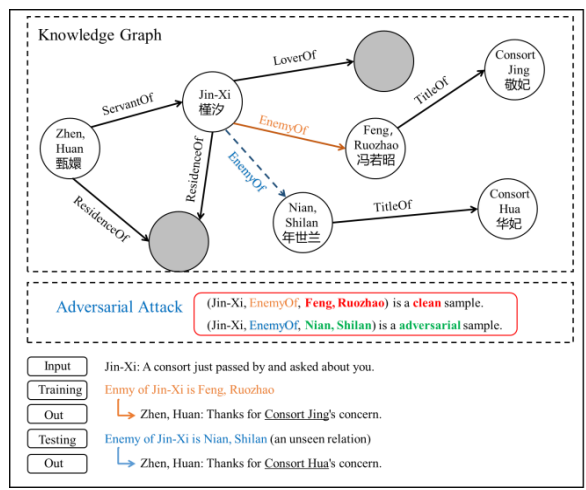

Figure 1: An example of an ideal conversation model with dynamic knowledge graphs and an example of adversarial attack.

To the best of our knowledge, it is the first work that combines dynamic knowledge graph with meta-learning to guide the conversation generation. The main contributions are as following:

1. Using meta-learning for knowledge dialogue tasks on a limited number of triples (each knowledge graph contains fewer entities and relationships), learning meta-parameters effectively to adapt to knowledge-aware dialogue system.

2. Studying how to quickly train a dynamic knowledge graph-based dialogue model (especially a seq2seq model) using a small dataset with both clean and adversarial samples as shown in Figure 1.

## 2. Related Work & Background

### 2.1 Knowledge Graph-based Conversations

Recently, there exist several models leveraging structured knowledge including factoid (Liu et al., 2018; Xu et al., 2017) or commonsense knowledge (Young et al., 2018; Zhou et al., 2018a) for generating appropriate and informative responses. In addition to presenting better dialogue systems, researchers also constructed and released several knowledge aware datasets (Tuan et al., 2019; Wu et al., 2019). Ghazvininejad et al. (2018) utilized memory network to store unstructured knowledge. (Zhou et al., 2018a) used knowledge graph embedding methods (e.g., TransE (Bordes et al., 2013)) to encode each knowledge triple. Liu et al. (2019) proposed a knowledge-aware dialogue system based on an augmented knowledge graph with both triples and texts. However, these works are somehow limited by small-scale graph or incomplete knowledge graph, and not adaptive to dynamic knowledge graph, where entities or relations are added or deleted. Compared with them, we propose a knowledge aware dialogue model, which fully consider the incompleteness of knowledge graphs and dynamic knowledge graphs.

## 2.2 Meta-Learning

Meta-learning or learning-to-learn aims at adapting quickly to new tasks with few steps and small datasets based on an optimal initialization. Recently, it has been applied on few-shot learning, such as image classification (Finn et al., 2017), machine translation (Gu et al., 2018), dialogue system (Qian and Yu, 2019; Lin et al., 2019), language generation (Huang et al., 2018), etc. There are three categories of meta-learning: 1. Metric-based (Koch et al., 2015; Vinyals et al., 2016; Santoro et al., 2016; Sung et al., 2018): learning a metric space and then comparing low-resource testing tasks to rich training tasks by using this space. 2. Policy-based (Andrychowicz et al., 2016; Munkhdalai and Yu, 2017; Mishra et al., 2017): learning a policy to update model parameters with few training tasks. 3. Optimization-based (Yin et al., 2018; Finn et al., 2017; Gu et al., 2018; Qian and Yu, 2019; Lin et al., 2019; Huang et al., 2018; Yoon et al., 2018): learning a model parameter initialization adapting quickly to new tasks. In this paper, we focus on learning an optimal conversation model initialization with improved adversarial meta-learning, which formulates dynamic knowledge triples as a problem of adversarial samples.

## 2.3 Adversary Attack

An adversarial sample refers to an instance with small, intentional feature perturbations that cause a learning model to make a false prediction. Currently, there are several studies of robustness to adversarial attacks (Goodfellow et al., 2015; Kurakin et al.,2017; Kurakin et al., 2017; Ilyas et al., 2019;Yuan et al., 2019; Zugner et al., 2019). Goodfellow et al. (2015) mainly explored the principle of adversarial sample attack and training using adversarial ideas. Kurakin et al. (2017) introduced the method of generating adversarial samples and adversarial training. Yuan et al. (2019) summarized the typical attacks, defense methods, and applications based on attack defense in this direction so far. However, we consider the dynamic changes between knowledge graphs as adversarial attacks.

## 3. Problem Formulation

For knowledge graph-based dialogue model $M_\theta$, the dialog context X and output response Y are paired with knowledge graph $\mathcal{G}$, which is composed of knowledge triples (h, r, t), where h, t are entities and r refers to relationship. The model $M_\theta$ is expected to generate a sentence that is not only similar to the ground-turth Y, but is consistent to the knowledge entities and relationships. When r or t are changed to r' or t', the generation sentence also correlates to the knowledge r' or t'.

The underlying idea of improved ADML is to utilize a set of tasks $\{\mathcal{T}_1,…, \mathcal{T}_K\}$ and find the model initialization adaptive to new task, and each task has a loss function $\mathcal{L}_i$ and contains a dataset $D_i$ (D = $\{(x_n, y_n, \mathcal{G}), n = 1…N\}$) that is further splited into clean and adversarial samples as $D_{clean_i}^{train}$, $D_{adv\ i}^{train}$, $D_{clean_i}^{test}$, $D_{adv\ i}^{test}$. Then, we compute loss on ($D_{clean_i}^{train}$, $D_{adv\ i}^{train}$), and perform gradient descent to find the optimal parameter as $\theta'_{clean_i}$ and $\theta'_{adv_i}$ respectively. In meta-update stage, we next find the optimal parameter θ depended on θ' obtained in the previous step.

## 4. KDAD: dynamic KG-based dialogue model with ADML

To our best knowledge, no prior knowledge-grounded conversation model utilized adversarial meta-learning algorithm to process incomplete knowledge triples and adapt to dynamic knowledge graph–based conversation. Our model is composed of (1) dynamic knowledge-grounded conversation module, and (2) an improved adversarial meta-learning module.

## 4.1 Knowledge Graph-based Dialogue Model

In knowledge graph-based dialogue system, $\mathcal{G} = \{H, R, T\}$ refers to a knowledge graph, where H, T $\in \mathcal{V}$ (the set of entities or attributes), R is a set of relationships, (h, r, t) is a knowledge triplets in KG. Given a input message $X = \{x_1, x_2,\ldots, x_m\}$ and $\mathcal{G}$, the goal is to generate a sequence $Y = \{y_1, y_2, \ldots, y_n\}$ with the knowledge aware dialogue systems. In general, the system consists of two stages: (1) knowledge selection: the model selects the entities $\mathcal{V}$ to maximize the following probability as candidates:

$$v_Y = \arg\max_V P(v | v_X, \mathcal{G}, X) \quad (1)$$

$v_X$ refers to one of entities retrieved from $\mathcal{G}$, which is connected to the entities or words in X. $v_Y$ refers to entities by performing knowledge graph reasoning to arrive at the vertex, which contains the knowledge for dialogue generation; (2) knowledge aware dialogue generation: it estimates the probability:

$$P(Y|X, v_Y) = \prod_{t=1}^{n} P(y_t | y_{<t}, X, v_Y) \quad (2)$$

The Qadpt model (Tuan et al., 2019) is constructed based on a seq2seq model incorporating knowledge graph reasoning, which is utilized to select knowledge consistent to dialogue content. Given an input context x, the encoder output a vector e(x), the decoder decode a vector $d_t$ based on the ground-truth or predicted y:

$$\begin{aligned} e(x) &= GRU(x_1 x_2 x_3 \ldots x_m) \\ d_t &= GRU(y_1 y_2 y_3 \ldots y_{t-1}, e(x)) \end{aligned} \quad (3)$$

where $d_t$ decides to copy knowledge graph entities or generic words, and generates the path matrix $R_t$ for knowledge graph reasoning, as shown in the following formula. These two processes are considered as knowledge selection module.

$$\begin{aligned} p(\{KB, W\} | y_1 y_2 y_3 \ldots y_{t-1}, e(x)) \\ = soft\max(\phi(d_t)) \end{aligned} \quad (4)$$

$$\begin{aligned} w_t &= p(W | y_1 y_2 y_3 \ldots y_{t-1}, e(x)) \\ c_t &= p(KB | y_1 y_2 y_3 \ldots y_{t-1}, e(x)) \\ o_t &= \{c_t k_t; w_t\}, \end{aligned} \quad (5)$$

where the probability $c_t$ refers to the controller which is used to choose entities V from knowledge graph, while the probability $1 - c_t$ is used to choose generic words W. φ is a fully connected neural network, and $k_t$ is the predicted distribution over knowledge graph entities V, and $o_t$ is the produced distribution over all vocabularies.

$$R_t = soft\max(\theta(d_t)) \quad (6)$$

$$A_{i,j,\gamma} = \begin{cases} 1, (h_i, r_j, t_\gamma) \in K \\ 0, (h_i, r_j, t_\gamma) \notin K \end{cases} \quad (7)$$

$$T_t = R_t A \quad (8)$$

where θ is a linear transformation operation, $R_t$ refers to the probability distribution of each head $h \in V$ choosing each relation type $r \in L$. $T_t$ is a transition matrix. A is an adjacency matrix which is a binary matrix indicating if the relations between two entities exist.

$$k_t = s^T (T_t)^N \quad (9)$$

where s is a binary vector used to indicate whether each knowledge entity exists in the input message x. s is multiplied by the transition matrix $T_t$ to produce $K_t$ which is a probability distribution over knowledge entities, where N refers to multi-hop reasoning.

## 4.2 Improved ADML for KG-based Dialogue Generation

In this section, we first introduce the key idea of the improved adversarial meta-learning, and then indicate how to combine the ADML with Qadpt model. ADML is able to learn the varying correlation between clean and adversarial samples to obtain a better and robust initialization of model parameters. The design philosophy of ADML (Yin et al., 2018) is shown in Figure 2.

For each task $\mathcal{T}_i$, in the inner gradient update process, ADML updates θ' to the direction of the adversarial subspace (purple color) as well as clean subspace (red color) to reach two points $\theta'_{adv_i}$ and $\theta'_{clean_i}$ respectively. Then in the meta-update stage, based on $\theta'_{adv_i}$ and $\theta'_{clean_i}$, ADML further optimizes θ' to reach the optimal point $\theta_i^*$, which is expected to fall into the intersection of two subspaces.

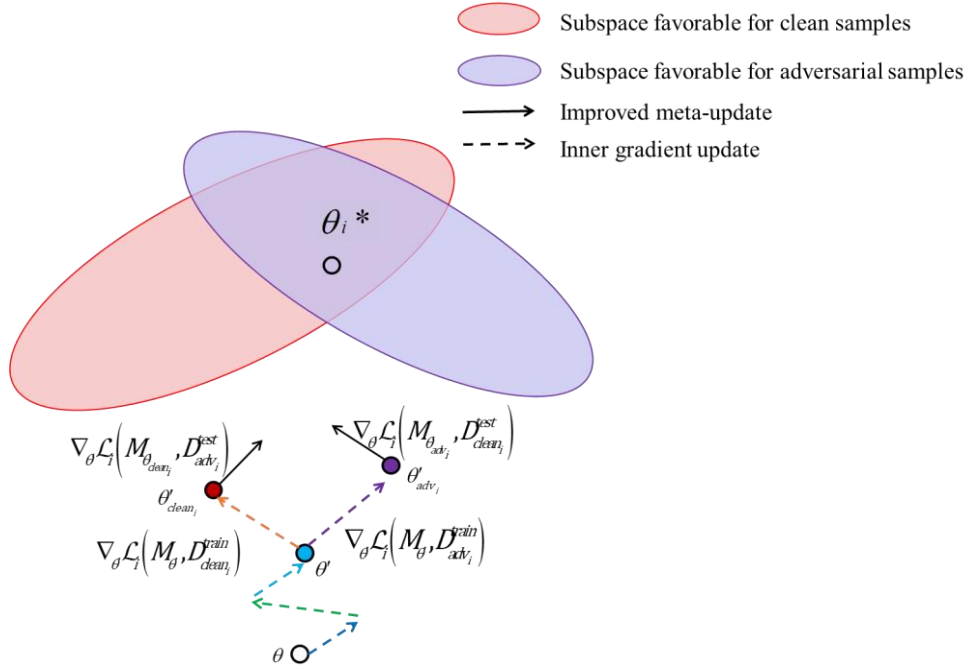

Figure 2: Illustration of design philosophy of improved ADML

We denote the knowledge aware model Qadpt as $M_\theta$ parameterized by θ, which is updated iteratively. At each step, we sample a batch of tasks $\{\mathcal{T}_1,…,\mathcal{T}_K\}$ containing support set $D_i^{train}$ and query set $D_i^{train}$ that are further splited into $D_{clean_i}^{train}$, $D_{adv_i}^{train}$. Then the model updates the parameters by k (k≥1) gradient descent steps for each task $\mathcal{T}_i$ in the following equations.

$$\theta_{clean_i}^k = \theta_{clean_i}^{k-1} - \alpha_1 \nabla_{\theta_{clean_i}^{k-1}} \mathcal{L}_{\mathcal{T}_i}\left(M_{\theta_{clean_i}^{k-1}}, D_{clearn_i}^{train}\right) \quad (10)$$

$$\theta_{adv_i}^{k+1} = \theta_{adv_i}^k - \alpha_2 \nabla_{\theta_{adv_i}^k} \mathcal{L}_{\mathcal{T}_i}\left(M_{\theta_{adv_i}^k}, D_{adv_i}^{train}\right) \quad (11)$$

$$\mathcal{L}_{\mathcal{T}_i}(\theta) = -\sum_{t=1}^n \log o_t(y_t) \quad (12)$$

where $\mathcal{L}_{\mathcal{T}_i}$ is the model loss function for task $\mathcal{T}_i$, $\alpha_1$ and $\alpha_2$ are the inner learning rate.

In the meta-update stage, we update the model parameters θ by optimizing the meta-objective function:

$$\min_\theta \sum_{\mathcal{T}_i \sim \mathcal{T}} \mathcal{L}_i\left(M_{\theta_{clean_i}^k}, D'_{adv_i}\right)$$
$$= \min_\theta \sum_{\mathcal{T}_i \sim \mathcal{T}} \mathcal{L}_i\left(M_{\theta_{clean_i}^{k-1} - \alpha_1 \nabla_{\theta_{clean_i}^{k-1}} \mathcal{L}_{\mathcal{T}_i}\left(M_{\theta_{clean_i}^{k-1}}, D_{clearn_i}^{train}\right)}, D'_{adv_i}\right) \quad (13)$$

$$\min_\theta \sum_{\mathcal{T}_i \sim \mathcal{T}} \mathcal{L}_i\left(M_{\theta_{adv_i}^{k+1}}, D'_{c_i}\right)$$
$$= \min_\theta \sum_{\mathcal{T}_i \sim \mathcal{T}} \mathcal{L}_i\left(M_{\theta_{adv_i}^k - \alpha_2 \nabla_{\theta_{adv_i}^k} \mathcal{L}_{\mathcal{T}_i}\left(M_{\theta_{adv_i}^k}, D_{adv_i}^{train}\right)}, D'_{c_i}\right) \quad (14)$$

The Meta-update of model $M_\theta$ is to update θ according to:

$$\theta = \theta - \beta_1 \nabla_\theta \sum_{\mathcal{T}_i \sim p(\mathcal{T})} \mathcal{L}_{\mathcal{T}_i}\left(M_{\theta'_{clean_i}}, D_{clean_i}^{train}\right) \quad (15)$$

$$\theta = \theta - \beta_2 \nabla_\theta \sum_{\mathcal{T}_i \sim p(\mathcal{T})} \mathcal{L}_{\mathcal{T}_i}\left(M_{\theta'_{adv_i}}, D_{adv_i}^{train}\right) \quad (16)$$

In our improved adversarial meta-learning algorithm, as shown in Figure 2, the meta-updated model parameter θ is the last step model parameter updated by num_task support sets. This parameter already contains information about the gradient direction of the support set instead of the random initial model parameter.

The overall learning process is shown in Algorithm 1, an updating episode includes an inner gradient update process (Line5–Line 9) and a meta-update process (Line 11).

---

**Algorithm 1** Training Algorithm of adversarial meta-learning model for knowledge graph-based dialogue generation ($M_\theta$, $\theta_0$, D, $\alpha_1$, $\alpha_2$, $\beta_1$, $\beta_2$)

---

**Input:** $M_\theta, \theta_0, D, \alpha_1, \alpha_2, \beta_1, \beta_2$
**Output:** $\theta^{Meta}$: optimal meta-learned model
1: Randomly initialize $\theta = \theta_0$
2: **while** not done **do**
3:     Sample batch of tasks $<\mathcal{T}_i>$ from task set T ;
4:     **for all** $\mathcal{T}_i$ **do**
5:        Sample K clean samples $\{(x_n^1, y_n^1, \mathcal{G}^1),\ldots,(x_n^K, y_n^K, \mathcal{G}^K)\}$ from $D_{clean_i}^{train}$;
6:        Sample K adversarial samples $\{(x_m^1, y_m^1, \mathcal{G}^1),\ldots,(x_m^K, y_m^K, \mathcal{G}^K)\}$ from $D_{adv_i}^{train}$ to form a dataset $D_i := \{D_{adv_i}, D_{clean_i}\}$ for the inner gradient update, containing K adversarial samples and K clean samples;
7:        Compute updated model parameters with gradient descent respectively:
        $\theta'_{adv_i} := \theta - \alpha_1 \nabla_\theta \mathcal{L}_i(M_\theta, D_{adv_i})$; $\theta'_{clean_i} := \theta - \alpha_2 \nabla_\theta \mathcal{L}_i(M_\theta, D_{clean_i})$;
8:        Sample K clean samples $\{(x_n^1, y_n^1, \mathcal{G}^1),\ldots,(x_n^k, y_n^k, \mathcal{G}^k)\}$ from $D_{clean_i}^{train}$;
9:        Sample K adversarial samples $\{(x_m^1, y_m^1, \mathcal{G}^1),\ldots,(x_m^k, y_m^k, \mathcal{G}^k)\}$ from $D_{adv_i}^{train}$ to form a dataset $D_i' := \{D_{adv_i}', D_{clean_i}'\}$ for the meta-gradient update, containing K adversarial samples and K clean samples;
10:    **end for**
11:    Meta Update $\theta := \theta - \beta_1 \nabla_\theta \sum_{\mathcal{T}_i \sim \mathcal{T}} \mathcal{L}_i(M_{\theta'_{adv_i}}, D'_{clean_i})$; $\theta := \theta - \beta_2 \nabla_\theta \sum_{\mathcal{T}_i \sim \mathcal{T}} \mathcal{L}_i(M_{\theta'_{clean_i}}, D'_{clean_i})$;
12: **end while**
13: Return $\theta^{Meta} = \theta$

---

## 5. Experiments

In this section, we first introduce the dataset and metric used to evaluate our model, and introduce the implementation details. Then, we describe our model evaluated in the experiments and compare our model with the strong baselines.

### 5.1 Dataset

For a fair comparison with the state-of-the-art knowledge aware dialogue model, Qadpt (Tuan et al., 2019) and TAware (Xing et al., 2017), we use the dataset, ("Hou Gong Zhen Huan Zhuang", HGZHZ), which first introduced to evaluate Qadpt. There are 174 KG entities and 9 KG relation types, 17164 dialog turns, 462647 tokens in HGZHZ. The dataset is split 5% as validation data and 10% as testing data. Because of the significant data imbalance of Friends, the result is often worse if Qadpt is tested directly. For imbalanced data, our model also cannot achieve satisfactory results. Imbalanced data will be the direction of our future research in dialogue system.

### 5.2 Evaluation Metrics

To verify whether our model can generate a more consistent and coherent response with reference to the given history dialogue and knowledge triples (even though model has not seen them), there are five main metrics in our experiments including BLEU, PPL, DISTINCT1/2/3/4 , KW/Genetric and Generated-KW to automatically evaluate the fluency, relevance, diversity, etc. The BLEU evaluates whether the generated response is also part of the task. PPL is a measurement of how well our model predicts a sample. DISTINCT measures the diversity of generated response. KW/Genetric and Generated-KW measures the capability of predicting the correct class (a knowledge graph entity or generic word) (Tuan et al., 2019).

### 5.3 Implementation Details

For all experiments, we set the learning rates $\alpha_1 = \alpha_2 = \beta_1 = \beta_2 = 0.001$, and these learning rates will adaptively decrease according to the loss. In our adversarial meta-learning stage, we set the num_task size to 4, support set size to 3, query set size to 4. We choose the one-layer GRU networks with a hidden size of 256 to construct the encoder and decoder. The model is optimized using Adam. In our model, we split data set to 4 buckets according the length of sentence, and then we utilize these buckets to train our model respectively.

### 5.4 Results and Analysis

Table 1 summarizes the experimental results of the proposed metrics for correctly predicting knowledge graph entities. We directly compare with the best results shown in (Tuan et al., 2019). We can observe that although TAware+multi method is overall better than the other models for KW/Generic, our model significantly outperforms other baselines on KW-Acc and Generated-KW. We also found that the methods using multi-hops reasoning technology outperform those without using multi-hops reasoning. From this table, it can be seen that our proposed generation model has better capabilities on knowledge graph entities prediction.

To evaluate the generated sentence quality, Table 2 presents the BLEU scores, perplexity (PPL) scores, and DISTINCT-N (DistN) scores. The results show that our model can achieve a high consistency score (BLEU), which is better than TAware, TAware+multi, Qadpt, and slightly less than Qadpt+multia. We can observe that our model significantly outperforms TAware, Qadpt, Qadpt +multi, in the perplexity (PPL). It can be seen that our method has significantly better performances in terms of distinct-n scores. We also found that the methods using multi-hop reasoning outperform those without using multi-hop reasoning, which confirms the benefits of using multi-hop reasoning in knowledge graph. In summary, our model can better control the generation to maintain its coherence, fluency, relevance, and diversity with the dialog history and knowledge graphs.

As shown in Table 3, we only utilize very small datasets to compare our model with the Qadpt model. We can see that our model achieves better results, which can prove that proposed method is robust to very small datasets. However, the compared model is severely overfitting and does not fit into small data sets.

|        | KWAcc | KW/Generic | | Generated-KW | |
|--------|-------|--------|-----------|--------|-----------|
|        |       | Recall | Precision | Recall | Precision |
| TAware | 50.21 | 44.40  | **35.50** | 49.18  | 76.72     |
| +multi | 57.71 | **68.61** | 28.70  | 44.50  | 90.70     |
| **Qadpt** | 57.61 | 38.24 | 28.31  | 44.50  | 90.70     |
| +multi | 57.40 | 51.97  | 28.43     | 44.50  | **91.22** |
| Our    | **59.37** | 40.37 | **34.15** | 47.82 | 90.07   |

Table 1: The results of knowledge graph entities prediction

| Models | BLEU | PPL | Dist1 | Dist2 | Dist3 | Dist4 |
|---|---|---|---|---|---|---|
| TAware | 14.14 | 90.11 | 0.011 | 0.061 | 0.135 | 0.198 |
| +multi | 13.34 | **80.48** | 0.022 | 0.122 | 0.122 | 0.239 |
| Qadpt | 14.52 | 88.24 | 0.013 | 0.081 | 0.169 | 0.242 |
| +multi | **15.47** | 86.65 | 0.021 | 0.129 | 0.259 | 0.342 |
| Our | 14.95 | 82.49 | **0.031** | **0.157** | **0.312** | **0.415** |

Table 2: The results of responses generation with BLEU, perplexity (PPL), distinct scores (1-gram to 4-gram)

| Samples | Method | PPL | BLEU | Dist1 | Dist2 | Dist3 | Dist4 |
|---|---|---|---|---|---|---|---|
| 300 | Qadpt | 15240.94 | 19.35 | 0.266 | 0.716 | 0.842 | 0.827 |
|  | Our | **1206.18** | 12.93 | 0.084 | 0.167 | 0.194 | 0.202 |

Table 3 The results of responses generation with BLEU, perplexity (PPL), distinct scores (1-gram to 4-gram) with very few datasets

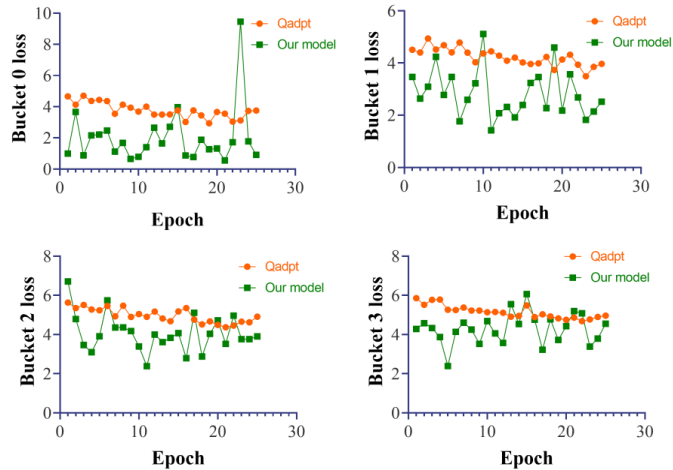

Figure 3: Loss curves of validation loss over epochs

Figure 3 shows the loss curves of evaluating our method and Qadpt on the validation set for different buckets. It can be observed from the figure that for different buckets, our four loss curves are almost all below Qadpt loss curves, which reflects the superiority of our model.

## 6 Conclusion

This paper proposes an algorithm for formulating dynamic knowledge graph as a problem of adversarial attack, focusing on the task of knowledge aware dialogue generation. We use adversarial meta-gradients to find the optimal initialization that is robust to changed KG path and can adapt to very small datasets. We achieve baseline results on HGZHZ comparing to several state-of-the-art models. Experimental results show that our knowledge graph-based dialogue generation model can make full use of knowledge triples to generate informative response.